\documentclass[letterpaper]{article} 
\usepackage{aaai24}  
\usepackage{times}  
\usepackage{helvet}  
\usepackage{courier}  
\usepackage[hyphens]{url}  
\usepackage{graphicx} 
\usepackage{natbib}  
\usepackage{caption} 
\usepackage{algorithm}
\usepackage{algorithmic}
\usepackage{amsmath}
\usepackage{comment}

\usepackage{newfloat}
\usepackage{listings}
\DeclareCaptionStyle{ruled}{labelfont=normalfont,labelsep=colon,strut=off} 
\floatstyle{ruled}
\newfloat{listing}{tb}{lst}{}
\floatname{listing}{Listing}
\title{AMSP-UOD: When Vortex Convolution and Stochastic Perturbation Meet Underwater Object Detection}

\author{
    Jingchun Zhou\textsuperscript{\rm 1}\equalcontrib,
    Zongxin He\textsuperscript{\rm 2}\equalcontrib,
    Kin-Man Lam\textsuperscript{\rm 3},
    Yudong Wang\textsuperscript{\rm 4},
    Weishi Zhang\textsuperscript{\rm 1},
    Chunle Guo\textsuperscript{\rm 5},
    Chongyi Li\textsuperscript{\rm 5}\thanks{Corresponding author}
}

\affiliations{
    \textsuperscript{\rm 1} School of Information Science and Technology, Dalian Maritime University\\
    \textsuperscript{\rm 2} School of Computer Science and Engineering, Huizhou University\\
    \textsuperscript{\rm 3} Department of Electrical and Electronic Engineering, Hong Kong Polytechnic University\\
    \textsuperscript{\rm 4} School of Electrical and Information Engineering, Tianjin University, China\\
    \textsuperscript{\rm 5} VCIP, CS, Nankai University\\
     zhoujingchun03@gmail.com, hikari0608@outlook.com, enkmlam@polyu.edu.hk, yudongwang@tju.edu.cn, teesiv@dlmu.edu.cn, \{guochunle, lichongyi\}@nankai.edu.cn
}

\usepackage{bibentry}

\begin{document}

\maketitle

\begin{abstract}
In this paper, we present a novel Amplitude-Modulated Stochastic Perturbation and Vortex Convolutional Network, AMSP-UOD, designed for underwater object detection. AMSP-UOD specifically addresses the impact of non-ideal imaging factors on detection accuracy in complex underwater environments. To mitigate the influence of noise on object detection performance, we propose AMSP Vortex Convolution (AMSP-VConv) to disrupt the noise distribution, enhance feature extraction capabilities, effectively reduce parameters, and improve network robustness. We design the Feature Association Decoupling Cross Stage Partial (FAD-CSP) module, which strengthens the association of long and short range features, improving the network performance in complex underwater environments. Additionally, our sophisticated post-processing method, based on Non-Maximum Suppression (NMS) with aspect-ratio similarity thresholds, optimizes detection in dense scenes, such as waterweed and schools of fish, improving object detection accuracy. Extensive experiments on the URPC and RUOD datasets demonstrate that our method outperforms existing state-of-the-art methods in terms of accuracy and noise immunity. AMSP-UOD proposes an innovative solution with the potential for real-world applications. Our code is available at: https://github.com/zhoujingchun03/AMSP-UOD.
\end{abstract}

\section{Introduction}
Recently, underwater object detection (UOD) has gained attention in the fields of marine technology, deep-sea exploration, and environmental protection. Precise detection of biological, geological, and man-made structures in deep-sea environments is vital for human society and environmental conservation \cite{xu2023systematic} \cite{zhuang2022underwater}. However, challenges in seawater, such as transparency, color, temperature, and suspended particles, combined with varying marine environments and target object types, reduce the accuracy of object detection.

Due to light absorption and scattering \cite{zhou2023camera,zhang2022underwaterimage}, underwater imaging often suffers from quality degradation compared to object detection in high-quality images. This impacts the performance of Convolutional Neural Network (CNN)-based object detectors. The key challenges include 1) the lack of underwater object detection datasets hindering the training of deep learning models, 2) degradation factors, such as light absorption and scattering, leading to low contrast and color distortion \cite{Zhou2023UGIF} \cite{guo2022underwater}, 3) the difficulty in extracting rich details from small and clustered underwater objects, and 4) class imbalance, making the challenging for object detectors to learn features for classes with a small-sample size \cite{fu2023rethinking}. To address these challenges, new detectors capable of accurate localization and classification in complex underwater environments are required. This research aims to advance ocean science and deep-sea exploration technology and holds practical value in environmental protection and resource development.

In this paper, we propose the AMSP-UOD network, crafted to tackle non-ideal imaging factors in underwater environments. Utilizing the optical imaging model $I=H(J,B,t)+N$ ($I$ represent observed image, $J$ represent raw scene, $B$ represents backscatter, and $t$ represent transmission map), we discern that underwater images combine a degradation function $H$ with noise $N$. To remove noise, we propose the AMSP-VConv. This strategy not only reduces parameters but also bolsters the network's robustness. We further implement the FAD-CSP to improve feature extraction in degraded environments. Our post-processing strategy, which relies on NMS, is designed to optimize the detection of dense clusters of underwater objects. Experimental results on the URPC \cite{urpc} and RUOD \cite{fu2023rethinking} datasets showcase the effectiveness of our method. Overall, AMSP-UOD presents an innovative solution for UOD with potential real-world applications.

The main contributions of this paper are as follows:

(1) We propose a novel single-stage UOD network. In the backbone, we design the AMSP-VConv to address the impact of noise and other degradations in underwater object detection. In the neck, the FAD-CSP boosts long and short distance feature connection, enhancing performance in complex underwater environments. Furthermore, an NMS-based post-processing method is introduced to enhance the detection performance of the network in complex underwater scenarios like dense waterweed clusters and fish schools.

(2) Our AMSP strategy refines network through parameter adjustments, enhancing detection by distinguishing between ideal and non-ideal imaging factors.

(3) Experimental results on public datasets and UOD competition datasets reveal that our method outperforms state-of-the-art UOD techniques in terms of both detection accuracy and speed. Ablation studies demonstrate that AMSP-VConv possesses superior noise resistance and interpretability, offering a novel solution for noise processing in detection tasks and computer vision.

\begin{figure*}[htbp]
\centering
\includegraphics[width=0.73\textwidth]{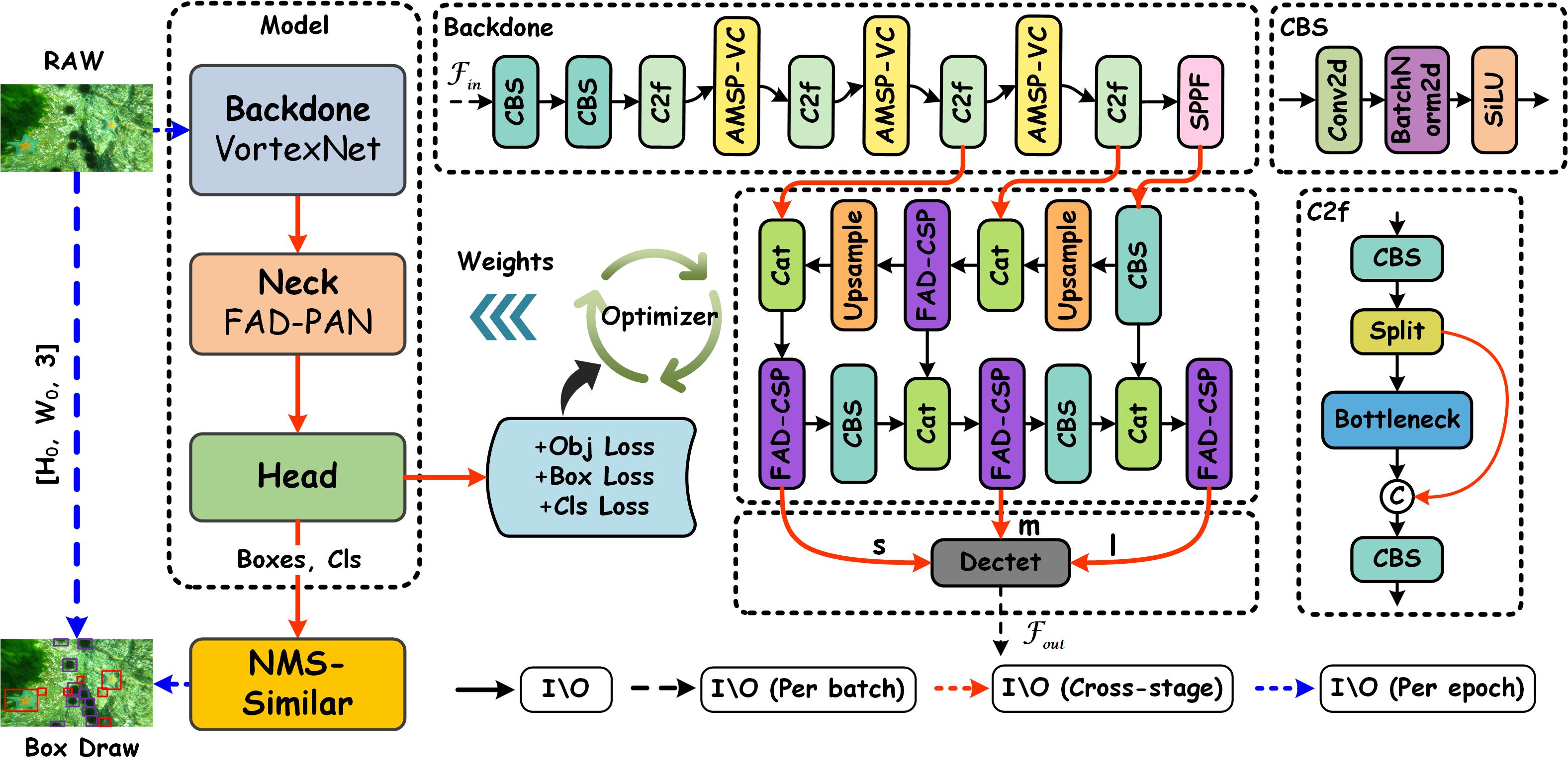}
\caption{AMSP-UOD network architecture: AMSP-VConv for underwater noise elimination; FAD-PAN for information analysis, FAD-CSP for semantic feature decoupling; NMS-Similar for merging traditional and Soft-NMS for efficient dense scene detection.}
\label{fig2}
\end{figure*}

\section{Related Work}
The UOD task focuses on detecting objects in underwater images. Deep learning has significantly advanced information fusion \cite{ma2023bilevel}, image enhancement \cite{liu2023holoco}, and object detection \cite{chen2022swipenet}. In many cases, these methods outperform traditional approaches, in terms of speed and accuracy \cite{liu2016ssd, ren2015faster, redmon2016you}. However, underwater environments introduce image degradations due to factors like light attenuation. Underwater robots also need efficient algorithms due to limited resources. Existing UOD techniques are either anchor-based or anchor-free, with variations in their approach.

\subsection{Anchor-Based Methods}
\subsubsection{Single-Stage Methods:} These methods predict the object's location and type directly, ensuring faster performance. Examples include SSD \cite{liu2016ssd} that leverages feature pyramids for multi-scale perception, RetinaNet \cite{lin2017focal} using Focal Loss for sample weight adjustment, and NAS-FPN \cite{ghiasi2019nas} that refines feature pyramid network structures. While efficient, they can struggle with precise object boundary localization in UOD tasks, especially in challenging conditions or with limited data. Data augmentation is often used to enhance generalization.

\subsubsection{Multi-Stage Methods:}
These techniques split detection into two stages: region proposal and object classification with bounding box prediction. Examples include Faster R-CNN \cite{ren2015faster}, Cascade R-CNN \cite{cai2018cascade}, DetectoRS \cite{qiao2020detectors}, and Dynamic R-CNN \cite{zhang2020dynamic}. They enhance accuracy using cascaded detectors, novel pyramid networks, and balanced learning \cite{ren2015faster, cai2018cascade, qiao2020detectors, zhang2020dynamic}. However, their computational demands pose challenges for on-the-go applications.

\subsection{Anchor-Free Methods}
\subsubsection{Key-Point Based Methods:} 
These techniques use key-points, either predefined or self-learned, for detection, offering finer object boundary detail. Examples are Reppoints \cite{yang2019reppoints} for learning object-related features, Grid \cite{fcos9010746} for grid-guided detection, and CenterNet \cite{zhou2019objects} and ExtremeNet \cite{zhou2019bottom} that use multiple key-points. While effective in general OD, their application in UOD is challenging due to limited underwater datasets \cite{fu2023rethinking}, manual annotations, and computational demands conflicting with UOD's typical scenarios.

\subsubsection{Center-Point Based Methods:}
These methods focus on predicting object center points, ideal for dense and fast detections. Notably, YOLO \cite{redmon2016you} approaches detection as a single regression task, optimizing dense object detection. Enhancements include per-pixel prediction and feature abstraction \cite{redmon2016you, fcos9010746, zhu2019feature, kong2020foveabox, liu2019high}. However, their scalability for various object sizes is limited, and they may not excel in tasks needing precise boundary localization, like specific underwater robot operations.

\section{Proposed AMSP-UOD Network}
The underwater environment is marked by complexity due to various regular and irregular degradation factors, including marine biological activity, human activity and current movement \cite{chou2021international}. These factors create unpredictable noise patterns, posing challenges to models attempting to perceive and model underwater degradation scenes. Underwater noise is complex compared \cite{li2019underwater} to typical noise conditions and requires a higher parameter count to denoise, but this increases the risk of overfitting. Instead of focusing on modeling noise, we propose a novel UOD network, namely ASMP-UOD (in Figure \ref{fig2}). Our approach aims to disrupt noise and reduce parameters, focusing on extracting ideal features rather than increasing the burden of noise analysis. Unlike previous methods that struggled with complex scenarios, ASMP-UOD is designed to better adapt to regular underwater scenes.

\begin{figure}[htbp]
\centering
\includegraphics[width=0.4\textwidth]{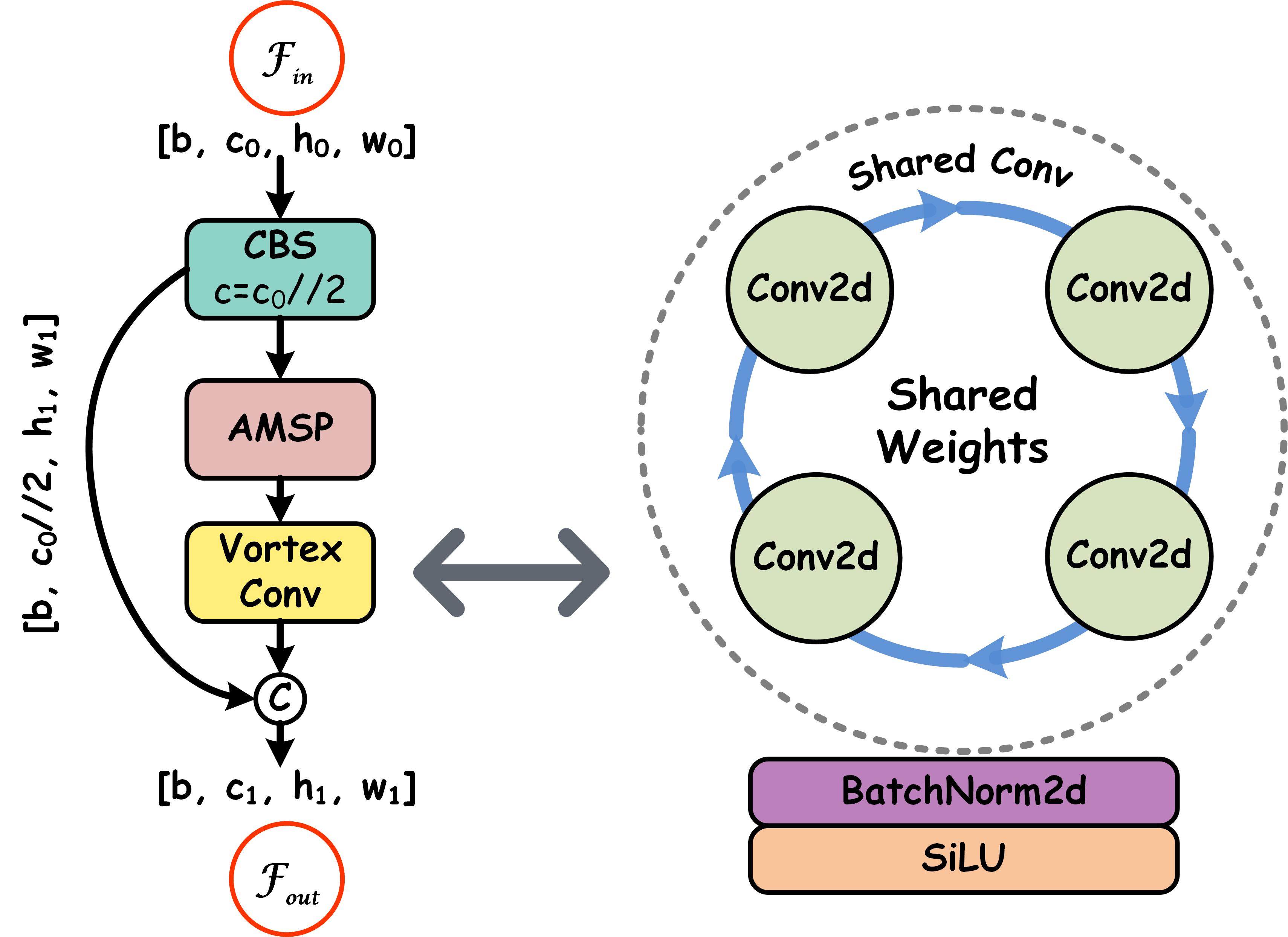}
\caption{AMSP Vortex Convolution. (a) The AMSP-VConv structure, (b) an expanded diagram of VConv, featuring the uniquely designed Shared Conv with BN, complemented by the SiLU activation function.}
\label{fig3}
\end{figure}

\subsection{Anti-Noise Capability of AMSP and VConv}
Convolution and its variants \cite{chollet2017xception}\cite{ghostnet} are crucial for feature extraction but often struggle in scenarios with noise interference or complex scenarios. The challenge lies in distinguishing between background features and target object features, limiting detection accuracy. To deal these issues, we design a novel AMSP-VConv to mitigate noise interference, enhancing the network's adaptability in underwater scenarios.

Inspired by the vortex phenomenon in turbulent water flows, which disrupts continuity through rapid rotation, AMSP-VConv introduces 'vortices' in the information flow to break the interference caused by noise. This innovation improves the network's ability to differentiate background and target features, enhancing detection in complex underwater environments.

In Figure \ref{fig3}, we present the complete structure of AMSP-VConv. Starting with an input tensor $F_{in}$ of the size $[b,c,h,w]$ ($b$: batch size, $c$: number of channels, $h$: height, $w$: width), it is processed by the combination of convolution, batch normalization, and the SiLU activation function (CBS) structure. This structure is designed to capture latent associations. By utilizing a kernel size of size 3 and a step size of 1, yielding an output tensor $X$ of size $[b,c//2,h,w]$. The transformation can be expressed as follows:
\begin{align}
X = CBS(F_{in}) = \delta (BatchNormal(Conv(F_{in}))) \label{cbs:silu}
\end{align}
where $\delta$ represents the SiLU activation function. As illustrated in Equations (\ref{amsp:1}) and (\ref{amsp:2}), we introduce the Amplitude Modulation and Shuffling Perturbation (AMSP) strategy in the subsequent steps. This strategy infuses random perturbations into the original grouped structure of associated features within $X$, thereby disrupting the association between noise and regular features. It is crucial to highlight that, while the AMSP strategy introduces these perturbations, it does not annihilate the features. Instead, it preserves a majority of the feature associations and induces a random shuffling among channels. This mechanism effectively serves the connection between noise samples and regular features, especially in the higher-level channels.
\begin{align}
T & = AM_{t}(X) = \begin{bmatrix}
    c_1 & c_2 & \ldots & c_t \\
    c_{t + 1} &  c_{t + 2} & \ldots & c_{2t} \\
    \ldots & \ldots & \ldots & \ldots \\
    c_{kt + 1} & c_{kt + 2} & \ldots & c_{kt + t}
\end{bmatrix} \label{amsp:1} \\ 
\notag \\
Y & = SP_t(T) = \begin{bmatrix}
    c_{a_0t + 1} &  c_{a_0t + 2} & \ldots & c_{a_0t + t} \\
    c_{a_1t + 1} & c_{a_1t + 2} & \ldots & c_{a_1t + t} \\
    \ldots & \ldots & \ldots & \ldots \\
    c_{a_kt + 1} & c_{a_kt + 2} & \ldots & c_{a_kt + t}
\end{bmatrix} \label{amsp:2} 
\end{align}
\begin{align}
\{a_0, a_1, \ldots, a_k\} = \{0, 1, \ldots, k\}
\end{align}
As depicted in Equations (\ref{amsp:1}) and (\ref{amsp:2}), the process involves two primary operations: Amplitude Modulation (AM) and Shuffling Perturbation (SP). AM is responsible for mapping the information to higher dimensions, SP perturbs these features. Here, we divide the channels into $k+1$ groups of $t$ channels each, $c_{i}$ denotes the $i$-th channel, and $Y$ is the output of the AMSP, which is aligned with the dimensions of the intermediate variable $T$.
\begin{align}
Z & = Concat(VConv(Y)) \\
Z^{'} & = \delta (BatchNormal(Z))
\end{align}

The VConv processes the reconstructed result $Z$ to optimize the extracted features. Drawing a parallel with the ideal state of water vortices, vortex convolution comprises multiple spiral lines (group convolutions) with a fixed spacing (shared convolution parameters). These group convolutions capture and extract features according to global and local imaging rules, removing isolated noise.
\begin{align}
F_{out} = Concat(X, Z^{'})
\end{align}

To ensure the integrity of the correct feature semantic information, we employ residual connections to concatenate the original associated features $X$ and $Z^{'}$ to obtain the final output $F_{out}$,  with shape $[b,c,h,w]$. This method can better adapt to feature attenuation and noise effects, thereby acquiring complete and correct features of the ideal degradation scenario under the guidance of the gradient optimizer.

\begin{figure}[tbp]
\centering
\includegraphics[width=\columnwidth]{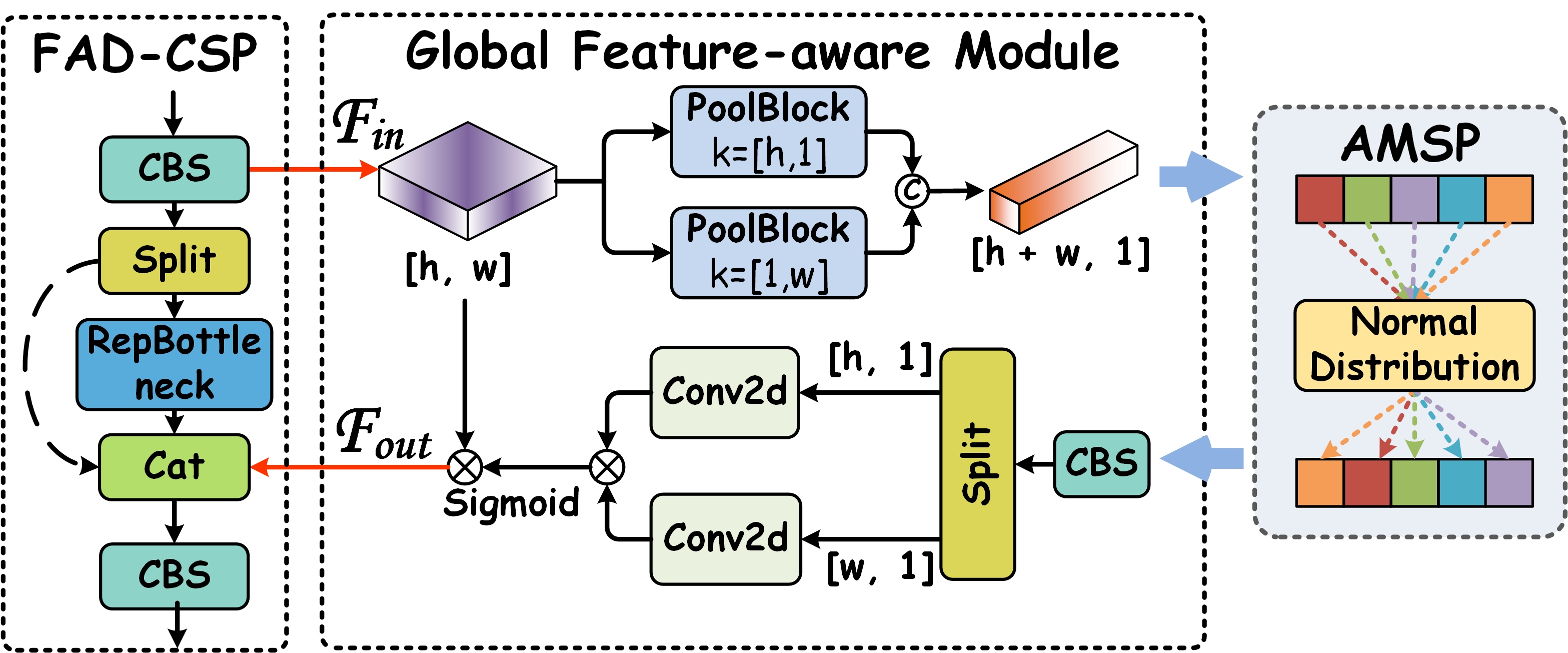}
\caption{FAD-CSP structure. The FAD-CSP module is built on a cross-stage network, comprising an efficient Global Feature-Aware (GFA) and a local decoupling-focused RepBottleneck. The essence of FAD-CSP lies in creating an efficient decoupling network through the interaction of long and short distance features.}
\label{fig4}
\end{figure}

\subsection{Feature Association Decoupling CSP}
In order to extract features at different distances for enhancing adaptability to underwater environments, we introduce a feature association and decoupling module based on a cross-stage network (FAD-CSP). This module is designed to incorporate the novel global feature-aware approach to extract long-range global features, while utilizing the optimized RepBottleneck as a sampling module to capture short-range local features.

\subsubsection{Global Feature-Aware Representation:}
In convolutional operations for global feature processing, a deeper network structure is usually required to extract rich feature information. This often increases the likelihood of the network getting trapped in local optima. To address this issue, we devised an efficient global feature-aware module and seamlessly integrated it into the FAD-CSP network using an attention mechanism. The structure of the proposed global feature-aware module is depicted in Figure \ref{fig4}. For a given input tensor $F_{in}$, we processing it through a bar-shaped pooling group, which compresses salient features into a one dimensional space. This method not only trades longer distance feature correlations, but also exhibits much lower computational overhead compared to convolution. This process can be expressed as follows:
\begin{figure}[htbp]
\centering
\includegraphics[width=0.4\textwidth]{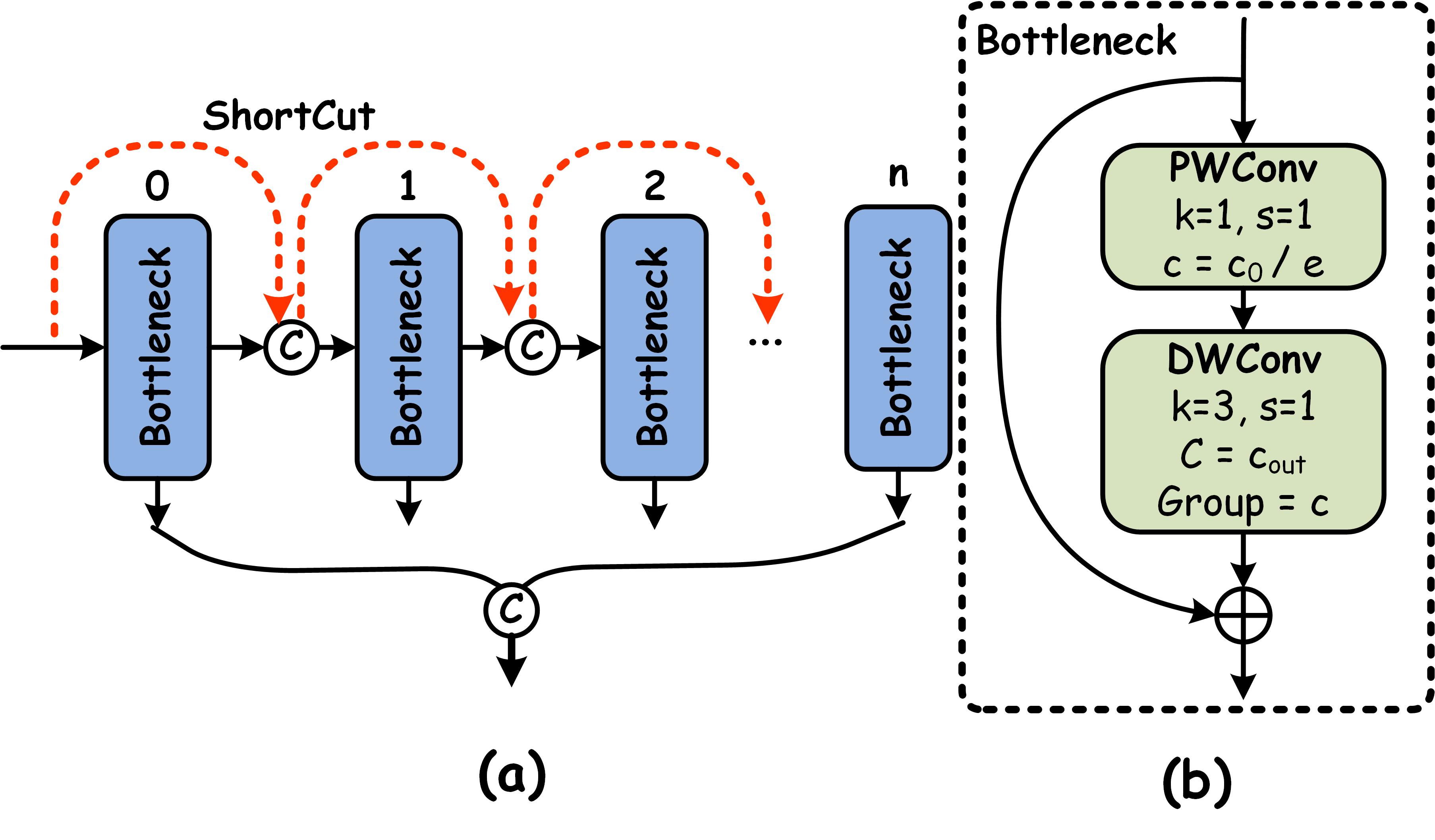}
\caption{(a) The represents the RepBottleneck structure with $n$=3, and (b) the detailed design of Bottleneck, a residual structure composed of pointwise convolution and depthwise convolution.}
\label{fig5}
\end{figure}

\begin{align}
v_c^h &= AvgPool_h(F_{in}) + MaxPool_h(F_{in}) \label{PoolGroup:1} \\
v_c^w &= AvgPool_w(F_{in}) + MaxPool_w(F_{in}) \label{PoolGroup:2}
\end{align}

The variables $v_h$ and $v_w$ are introduced into the subsequent stage of global feature-aware processing. Utilizing the AMSP strategy, they undergo a random alternation. This procedure generates a distribution map highlighting global prominent features as follows:
\begin{align}
&y_{f} = CBS(AMSP(Cat_c([v_c^h, v_c^w]))) \\
&y_{c/r}^h, y_{c/r}^w = split_{c}(y_{f})
\end{align}
where $c$ denotes the channel count of this intermediary value and $r$ is the scaling ratio. The CBS is employed and $Cat_c$ denotes concatenation by channels to rebuild feature relationships $y_{f}$, extracting accurate long-distance feature associations.
\begin{align}
& y_c^h = Conv(y_{c/r}^h), y_c^w = Conv(y_{c/r}^w)
\end{align}
 
To capture long-range features from the attention map, we incorporate a weighted gradient flow. This ensures the retention of valuable information in the output and upholds the consistency within the original feasible solution domain. Specifically, the adaptive weighting stems from redistributing two distinct linear feature sets sourced from independent conventional mappings and harnessing the capabilities of the Sigmoid function. This method allows for adaptive weighting of the features, according to the disparities in feature importance within specific regions, ensuring a refined adjustment to feature changes across different areas.
\begin{align}
& A_f = Sigmoid(y_c^h \times  y_c^w) \\
& F_{out} = A_f \bigodot  F_{in} 
\end{align}

Ultimately, the global attention $A_f$, obtained by weighting the product of the strip attention maps $ y_c^h \times  y_c^w$ followed by the Sigmoid function, is multiplied with the input $F_{in}$. This generates an expanded solution domain $F_{out}$, reinterpreted by the global feature perception decoupling module. It provides the network with a richer and optimized feature representation. Introducing the attention map allows the network to better understand and process features from different regions while retaining key information. This aids the network in achieving global optima, enhancing the performance of the UOD task.

\subsubsection{RepBottleneck:} This is an efficient residual structure, as shown in Figure \ref{fig5} (b), which uses a combination of depthwise separable convolutions and residual connections to reduce the number of network parameters, aggregating local features within the receptive field. Our RepBottleneck is an optimization over the Bottleneck List, addressing the deficiency in global feature-awareness. RepBottleneck focuses on local representations at a short distance. The proposed RepBottleneck is depicted in Figure \ref{fig5}(a), which interconnects multiple Bottlenecks and ShortCuts, to enhance the degree of association between local features. It is expressed as follows:
\begin{align}
I_{n} = 
\begin{cases}
Bottleneck((I_{0}), & if\text{ } n = 1 \\
Bottleneck(Cat(I_{n-1}, I_{n-2})), & if\text{ } n \neq 1 \\
\end{cases}
\end{align}
where $I_{n}$ represent the output of RepBottleneck and $I_{0}$ represent the input of RepBottleneck. 

Eventually, FAD-CSP obtains rich local features, abstract global features, and separated primitive features. As shown in Figure 
\ref{fig4}, FAD-CSP uses CBS to associate long and short distance features related to the target, decoupling irrelevant degraded features and realizing the improvement of detection accuracy.

\subsection{Non-Maximum Suppression-Similar}
In dense underwater environments, two primary challenges in detection are overlapping objects with similar features and overlapping bounding boxes for the same target, leading to inaccuracies in traditional NMS methods. While Soft-NMS \cite{bodla2017soft} retains more boxes, it increases computational time. To overcome these issues, we propose an NMS method based on aspect ratio similarity, called NMS-Similar. This method combines traditional NMS's speed with SoftNMS's precision, using a unique aspect ratio threshold and optimized greedy strategy. The suppression mechanism for each object is as follows:
\begin{align}
& S_{i} = S_i e^{-IoU(M, b_i)^2 /\sigma} \\
& L' = (IoU(b_i, L) <= N_t) \text{ and } (Sim(M, L) > N_s) \label{cal_times} \\
& Sim(M, b_i) = \frac{\vec{M} \cdot \vec{b_i}}{\|\vec{M}\| \|\vec{b_i}\|}
\end{align}
where $S_i$ is the current detection box's confidence, $M$ is the highest confident box, Intersection over Union (IoU) measures overlap between predicted and ground-truth boxes, $N_t$ is the preset IoU threshold, $Sim$ calculates the aspect ratio similarity, \( \vec{M} \) represents the length and width of $M$. $N_s$ is the preset similarity threshold, and $\sigma$ is a Gaussian weighting function. $L$ and $L'$ represent the remaining and recalculated detection boxes, respectively. Equation (\ref{cal_times}) adjusts the suppression counts for non-maximum confidence boxes by introducing an aspect ratio threshold to exclude similar detection boxes. The threshold strategy takes into account that object detection boxes for the same object at different scales share similar aspect ratios. During the computation process, similar detection boxes are precluded in advance, reducing the suppression time in dense scenes while ensuring detection accuracy.

\section{Experimental Results}
We elaborate on our experimental setup and comparative analyses. Experiments reveal that our approach significantly enhances the network's accuracy and resistance to noise, especially in challenging underwater conditions.

\subsection{Implementation Details}
Our experiments run on an Intel Xeon E5-2650 v4 @ 2.20G CPU and an Nvidia Tesla V100-PCIE-16GB GPU with the Ubuntu 20.04 LTS operating system and Python 3.10 environment built on Anaconda, with a network architecture based on Pytorch 2.0.1 build. The hyperparameters are shown in Table \ref{tab:1}.

In addition, if not specified, the comparison experiments are performed using the traditional NMS method. 
\begin{table}[htbp]
\centering
\fontsize{10pt}{13pt}\selectfont
\begin{tabular}{cc|cc}
\hline
Type & Setting & Type & Setting \\
\hline
Image size & 640 & Weights & None \\
Batch-size & 16 & Seeds & 0\\
Optimizer & SGD & LR & 0.01\\
Epochs & 300 & Early-stop & True \\
\hline
\end{tabular}
\caption{Hyperparameter settings}
\label{tab:1}
\end{table}
\fontfamily{\familydefault}\selectfont

\begin{figure*}[tbp]
\centering
\includegraphics[width=0.85\textwidth]{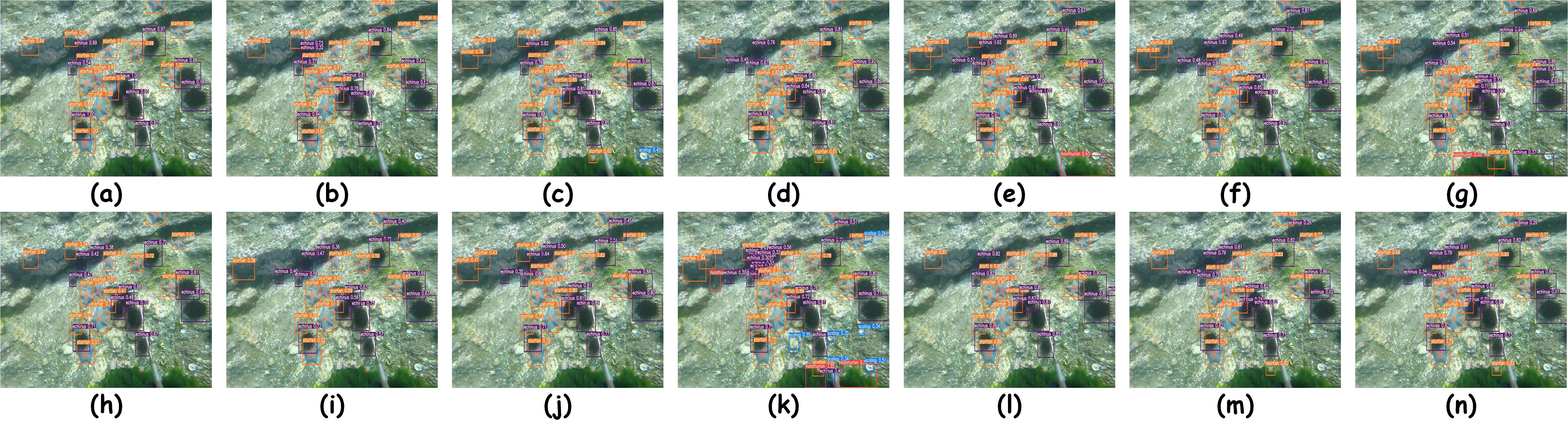}
\caption{Visualization of object detection results of different object detection methods on URPC (Zhanjiang). (a) YOLOv3 \cite{redmon2018yolov3}, (b) YOLOv5s \cite{yolov5}, (c) YOLOv6s\cite{li2022yolov6}, (d) YOLOv7-tiny \cite{wang2023yolov7}, (e) Faster R-CNN \cite{girshick2015fast}, (f) Cascade R-CNN \cite{cai2018cascade}, (g) RetinaNet \cite{lin2017focal}, (h) FCOS \cite{fcos9010746}, (i) ATSS \cite{zhang2020bridging}, (j) TOOD \cite{feng2021tood}, (k) PAA \cite{kim2020probabilistic}, (l) Ours-Standard, (m) Ours-AMSP-VConv, (n) Ours-AMSP-VConv + NMS Similar.}
\label{fig6}
\end{figure*}

\subsection{Evaluation Metrics and Datasets}
We adopt AP and AP50 as the primary metrics for model accuracy evaluation, with precision (P) and recall (R) as supplementary indicators. To showcase the generalizability of our network, we trained it on the URPC (Zhanjiang) \cite{urpc} dataset, from the 2020 National Underwater Robotics Professional Competition and the extensive RUOD dataset. The URPC dataset contains 5,543 training images across five categories, with 1,200 images from its B-list answers serving as the test set. The RUOD dataset \cite{fu2023rethinking} contains various underwater scenarios and consists of 10 categories. It includes 9,800 training images and 4,200 test images.

\subsection{Visual Comparisons}
Figure \ref{fig6} visualizes the object detection results of different detection frameworks on the URPC (Zhanjiang) dataset. Many of these frameworks struggle to accurately detect smaller objects, with some even mistakenly identifying the background as a target. The Faster R-CNN \cite{ren2015faster}, RetinaNet \cite{lin2017focal}, and PAA methods exhibit false positives by detecting kelp as seagrass. In contrast, the YOLO methods \cite{redmon2016you} miss some objects, failing to detect certain starfish. Our method excels in detecting smaller objects without any false positives or missed detections.

\begin{table}[!h]
\centering
\setlength{\tabcolsep}{2pt}
\fontsize{10pt}{13pt}\selectfont
\begin{tabular}{cccccc}
\hline
Baselines & Time & Memory & P & R & $mAP_{0.5}^{0.95}$\\
\hline
DSC & 15s & \textbf{4.61G} & 0.824 &  0.510 & 0.371 \\
GC  & 52s & 5.10G & 0.796 & 0.637 & 0.396\\
VC (w/o SW) & 14s & 5.36G & 0.730 & \textbf{0.694} & 0.386 \\
VC (w/o AMSP) & 14s & 4.65G &  0.833 & 0.631 & 0.397\\
AMSP-VConv & \textbf{14s} & 4.65G & \textbf{0.845} & 0.612 & \textbf{0.398}\\
\hline
\end{tabular}
\caption{Ablation of AMSP-VConv. Time: Inference Time (per epoch), DSC: Depthwise Separable Conv, GC: Ghost Conv, SW: Shared Weight, VC: AMSP-VConv, P: Precision, R: Recall.}
\label{tab:4}
\end{table}

\subsection{Quantitative Comparisons}
In Table \ref{tab:3}, the performance of various versions of AMSP-UOD on the URPC and RUOD datasets is presented. Notably, while our AMSP-VConv version indicates slightly reduced stability and precision compared to the Our-Standard version in balanced scenarios, it showcases enhanced detection capability in more degenerative conditions (URPC). This observation is also substantiated by subsequent ablation studies. We believe this significant improvement can be attributed to the noise suppression capability of the VConv design combined with the outstanding feature perception ability of FAD-CSP. Especially in intricate underwater environments, our method adeptly boosts the recognition accuracy of waterweeds, which are treated as a small-sample target, to a remarkable 99.3\%. Furthermore, the integration of the NMS-Similar strategy imparts a clear enhancement in detection rates for the Vortex version. This strategy efficiently curtails false positives and misses, thus ensuring the integrity and accuracy of object detection. In comparison with the series of YOLO models and other leading detection techniques, our method consistently manifests marked superiority on a foundation of high precision. In conclusion, our method exhibits exemplary efficiency and adaptability in UOD, underscoring its profound potential for real-world underwater applications.

\begin{table*}[htbp]
\centering
\fontsize{10pt}{13pt}\selectfont
\begin{tabular}{c|cc|ccccc|ccc}
\hline
\multicolumn{1}{c|}{\raisebox{-0.5\normalbaselineskip}{Method}} & \multicolumn{2}{c|}{URPC} & \multicolumn{5}{c|}{URPC Categories AP50 } & \multicolumn{2}{c}{RUOD} \\
\cline{2-10}
 & AP↑ & $AP_{50}$↑ & Ho↑ & Ec↑ & St↑ & Sc↑ & Wa↑ & AP↑ & $AP_{50}$↑\\ 
\hline  
YOLOv3 & 29.7 & 58.9 & 63.5 & 83.1 & 68.1 & 46.4 & 33.2 & 49.1 & 80.3  \\
YOLOv5s & 38.6 & 66.2 & 67.3 & 84.7 & \underline{76.7} & 57.2 & 43.0 & 53.8 & 81.4 \\
YOLOv6s & 36.1 & 62.8 & 61.4 & 85.2 & 68.1 & 49.0 & 50.1 & 60.1 &	84.9 \\
YOLOv7-tiny & 35.9 & 62.2 & 57.9 & 84.9 & 72.3 & 50.1 & 66.3 & 57.9 & 84.3 \\
\hline
Faster R-CNN & 31.0 & 59.0 & 66.9 & 85.9 & 72.1 & 55.4 & 14.7 & 49.1 & 80.3 \\
Cascade R-CNN & 31.6 & 59.1 & 67.1 & 86.0 & 71.3 & 56.2 & 14.7 & 53.8 & 81.4\\
RetinaNet & 26.3 & 51.1 & 61.3 & 81.8 & 66.2 & 46.2 & 0.00 & 48.0 &  77.8 \\
FCOS & 29.2 & 58.1 & 61.8 & 83.5 & 68.8 & 53.9 & 22.3 & 49.1 & 80.3 \\
ATSS & 29.0 & 55.6 & 64.0 & 84.8 & 71.4 & 55.8 & 2.20 & 53.9 &  82.2 \\
TOOD & 30.1 & 56.7 & 65.0 & 86.1 & 72.7 & \underline{58.3} & 1.30 & 55.3 & 83.1 \\
PAA & 34.2 & 62.3 & 65.1 & 85.2 & 70.9 & 55.9 & 34.6 & 53.5 & 82.2 \\
\hline
Ours (Standard) & \textbf{\underline{45.0}} & 73.4 &  \textbf{\underline{69.1}}  & 86.6 & 75.3 & 53.1 & 83.0 & \underline{62.1} & \underline{85.9} \\
Ours (AMSP-VConv) & 36.6 & \underline{74.8} & 62.9 & \underline{87.1} & 72.9 & 51.6 & \underline{99.3} & 61.4 & 85.3  \\
Ours (AMSP-VConv + NMS-Similar) & \underline{40.1} & \textbf{\underline{78.5}} & \underline{67.3} & \textbf{\underline{87.5}}  & \textbf{\underline{77.5}} & \textbf{\underline{60.6}} & \textbf{\underline{99.5}} & \textbf{\underline{65.2}} & \textbf{\underline{86.1}}  \\
\hline
\end{tabular}
\caption{Comparison with existing methods on the URPC and RUOD datasets. Ho: holothurian's AP50, Ec: echinus's AP50, St: starfish's AP50, Sc: scallop's AP50, Wa: waterweeds's AP50. AP: AP@[0.5:0.05:0.95], AP50: AP@0.5. Bolding and underlining is highest, underlining only is second-highest.}
\label{tab:3}
\end{table*}

\begin{table}[htbp]
\centering
\fontsize{10pt}{13pt}\selectfont
\resizebox{\columnwidth}{!}{
\begin{tabular}{cccccc}
\hline
Baselines & Time(ms) & $mAP_{0.5}$ & $mAP_{0.5}^{0.95}$ & $AP_{echinus}$ \\
\hline
NMS & \textbf{14.20} & 0.748 & 0.366 & 0.477 \\
Soft-NMS & 337.3 & 0.785 & 0.400 & 0.509\\
NMS-Similar & 46.90 & \textbf{0.785} & \textbf{0.401} & \textbf{0.509}\\
\hline
\end{tabular}}
\caption{Ablation of NMS-Similar}
\label{tab:5}
\end{table}
\fontfamily{\familydefault}\selectfont
\begin{table}[!h]
\centering
\fontsize{10pt}{13pt}\selectfont
\begin{tabular}{ccccc}
\hline
Baselines & P & R & $mAP_{0.5}$ & $mAP_{0.5}^{0.95}$ \\
\hline
a & 0.836 & 0.610 & 0.675 & \textbf{0.397} \\
b & 0.734 & 0.625 & 0.640 & 0.370 \\
c & 0.720 & 0.658 & 0.679 & 0.377 \\
d & \textbf{0.858} & 0.612 & 0.681 & 0.369 \\
All & 0.844 & \textbf{0.681} & \textbf{0.748} & 0.366 \\
\hline
\end{tabular}
\caption{Ablation of FAD-CSP}
\label{tab:6}
\end{table}
\fontfamily{\familydefault}\selectfont
\subsection{Ablation Studies}
In order to verify the impact of the proposed module on the network performance, we conducted a series of ablation experiments.

\subsubsection{Ablation of AMSP-VConv: }
In Table \ref{tab:4}, we find that the combination of VConv with the AMSP strategy provides an optimal balance, in terms of precision, recall, and mAP, while maintaining reasonable inference time and memory usage. Compared to Depthwise Separable Convolution (DSC) and Ghost Convolution (GC), AMSP-VConv demonstrates superior performance in complex object detection tasks, particularly in intricate scenarios that need to balance multiple performance metrics. The ablation experiments further reveal the importance of shared parameters and the AMSP strategy for enhancing both accuracy and efficiency. Ultimately, the integration of VConv and the AMSP strategy proves its potential in improving object detection tasks, providing robust support for real-world applications.

\begin{figure}[!ht]
\centering
\includegraphics[width=0.36\textwidth]{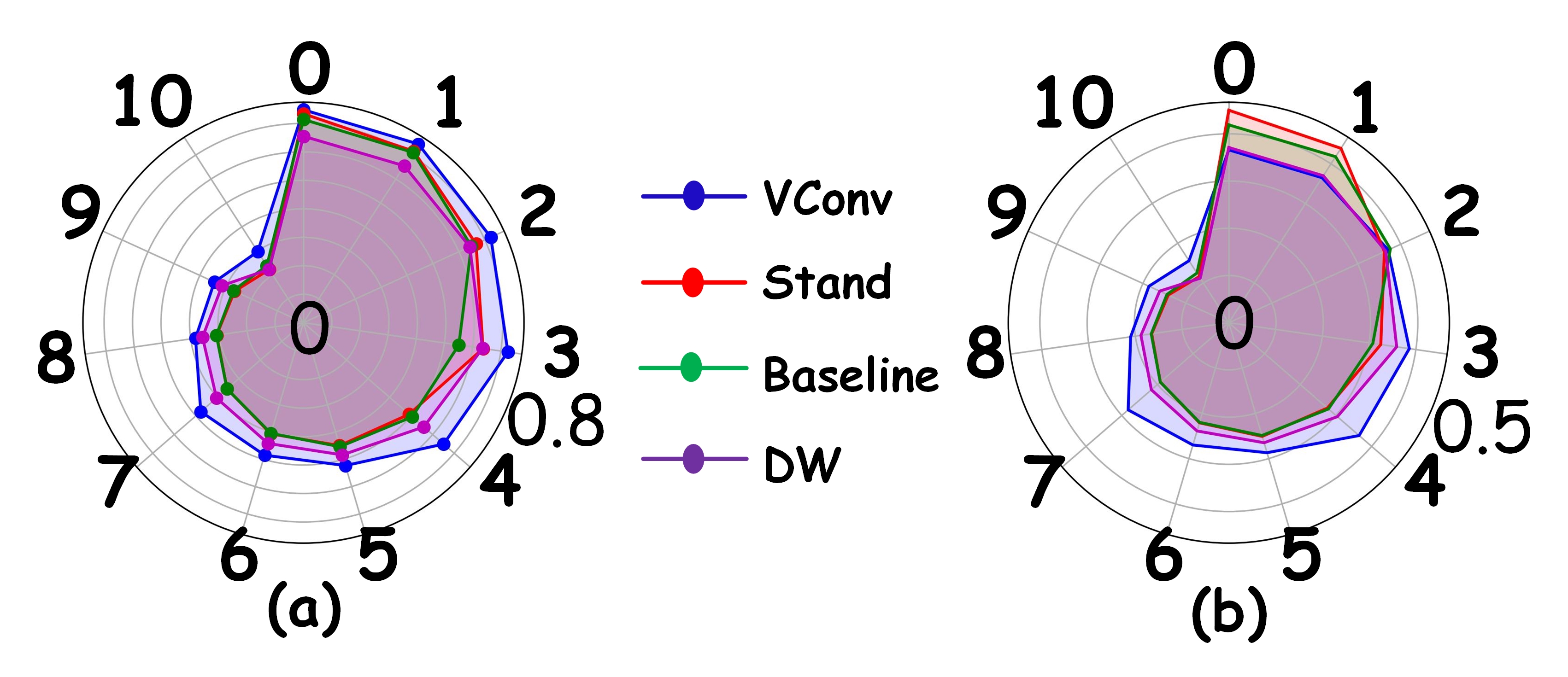}
\caption{Noise robustness ablation for AMSP-VConv. (a) and (b) show mAP@0.5 and mAP@0.5:0.95 under varied noise levels. Numbers 0-10 represent noise (0 + Gaussian noise standard deviation). Level 0 represents original underwater scene. Methods are not pre-trained on noisy images. Blue is AMSP-VConv, red is Standard Conv, green is YOLOv5s model, and purple is depthwise-separated Conv.}
\label{fig7}
\end{figure}

Underwater scenarios are susceptible to noise interference, and noise robustness is a crucial metric for evaluating UOD methods. Ensuring all operations that influence network metrics are equivalent, Gaussian noise was used to simulate the underwater noise environment, creating multiple noise levels (i.e., the original scenario augmented with Gaussian noise of varying standard deviations). We trained our network using the URPC dataset. As shown in Figure \ref{fig7}, our network's mAP score remains stable under the influence of noise level 4. In contrast, the mAP@0.5 of YOLOv5s, serving as the Baseline, decreased by 16.3\%. In high-noise scenarios, our AMSP-VConv demonstrates superior noise robustness, while the accuracy of Our-Standard, which merely replaces the AMSP-VConv module with standard convolutions, aligns closely with that of the Baseline. This indicates that AMSP-VConv in the Backbone network provides AMSP-UOD with strong noise robustness, validating the effectiveness of AMSP-VConv. It offers an excellent solution for denoising and complex underwater scenarios.

\subsubsection{Ablation of NMS-Similar: }
From Table \ref{tab:5}, it is evident that NMS-Similar achieves a commendable balance between accuracy and efficiency. Compared to Soft-NMS, NMS-Similar retains similar detection accuracy while significantly reducing computation time. Especially in challenging underwater detection scenarios, where closely located or overlapping objects are frequent, the performance of NMS-Similar stands out, underscoring its immense value in real-world applications.

\subsubsection{Ablation of FAD-CSP: }
In Table \ref{tab:6}, we evaluated the contributions of various components in the FAD-CSP. Four configurations are tested:
a) Without the GFA module. b) Replacing the Pooling Groups with Individual Pooling Layers. c) Removing the AMSP strategy from the GFA module in FAD-CSP. d) Replacing Repbottleneck with Bottleneck.
Among the tested configurations, using the FAD-CSP method achieves the best results, with the highest mAP of 0.748 and an improved recall rate of 0.681. This underscores the importance of each component in enhancing the detection performance. In particular, removing the GFA module (a) or the AMSP strategy from GFD (c) leads to a decrease in performance, highlighting their critical roles in the framework. Additionally, using Repbottleneck (as opposed to the standard Bottleneck) further bolsters the detection results, emphasizing its effectiveness in the context of the FAD-CSP method.

\section{Conclusion}
In this work, we proposed AMSP-UOD, a novel network for underwater object detection, addressing non-ideal imaging factors in complex underwater environments. With our innovative AMSP Vortex Convolution, we enhance feature extraction and network robustness, while our FAD-CSP module improves performance in intricate underwater scenarios. Our method optimizes detection in object-dense areas and outperforms existing state-of-the-art methods on the URPC and RUOD datasets. The practical evaluations highlight the potential applicability of AMSP-UOD to real-world underwater tasks, making it a promising contribution to UOD.

\section{Acknowledgments}
This work was supported in part by the National Natural Science Foundation of China (No.62301105), the 2022 National Undergraduate Innovation and Entrepreneurship Training Program Project (No.202210577003), National Key Research and Development Program of China (No.2018AAA0100400), China Postdoctoral Science Foundation (No.2021M701780), the High Performance Computing Center of Dalian Maritime University, and the Supercomputing Center of Nankai University. We are also sponsored by CAAI-Huawei MindSpore Open Fund.

\bibliography{aaai}

\end{document}